\documentclass{article}

 \usepackage[preprint]{neurips_2024}

\usepackage[utf8]{inputenc}
\usepackage[T1]{fontenc}
\usepackage{hyperref}
\usepackage{url}
\usepackage{booktabs}
\usepackage{amsfonts}
\usepackage{nicefrac}
\usepackage{microtype}
\usepackage{xcolor}
\usepackage{graphicx}
\usepackage{enumitem}

\title{Agyn: A Multi-Agent System for Team-Based Autonomous Software Engineering }

\author{
 Nikita Benkovich \\
  Agyn; Mila – Quebec AI Institute (e-Lab) \\
  Tel-Aviv, Israel \\
  \texttt{benkovich@agyn.io}
  \And
  Vitalii Valkov \\
  Agyn; Mila – Quebec AI Institute (e-Lab) \\
  Tel-Aviv, Israel \\
  \texttt{vitalii@agyn.io}
}

\begin{document}

\maketitle

\begin{abstract}
Large language models have demonstrated strong capabilities in individual software engineering tasks, yet most autonomous systems still treat issue resolution as a monolithic or pipeline-based process. In contrast, real-world software development is organized as a collaborative activity carried out by teams following shared methodologies, with clear role separation, communication, and review. In this work, we present a fully automated multi-agent system that explicitly models software engineering as an organizational process, replicating the structure of an engineering team. Built on top of \textsc{agyn}, an open-source platform for configuring agent teams, our system assigns specialized agents to roles such as coordination, research, implementation, and review, provides them with isolated sandboxes for experimentation, and enables structured communication. The system follows a defined development methodology for working on issues, including analysis, task specification, pull request creation, and iterative review, and operates without any human intervention. Importantly, the system was designed for real production use and was not tuned for SWE-bench. When evaluated post hoc on SWE-bench 500, it resolves 72.2\% of tasks, outperforming single-agent baselines using comparable language models. Our results suggest that replicating team structure, methodology, and communication is a powerful paradigm for autonomous software engineering, and that future progress may depend as much on organizational design and agent infrastructure as on model improvements.
\end{abstract}

\section{Introduction}

Large Language Models (LLMs) have recently demonstrated strong capabilities in software engineering tasks such as code understanding, bug fixing, and test generation. Benchmarks such as SWE-bench \citep{jimenez2024swebench} have emerged as a standard evaluation framework for measuring progress toward autonomous software engineering agents by requiring systems to resolve real GitHub issues through patches that pass repository test suites. Despite steady improvements in model quality, building agents that can reliably solve such tasks in a fully automated manner remains a significant challenge.

Most existing approaches to SWE-bench rely on single-agent architectures, where one agent is responsible for issue understanding, repository exploration, code modification, and validation. These agents are typically augmented with tools such as shell access, code editors, and search utilities, and operate within a fixed scaffold. While effective in many cases, this paradigm places heterogeneous responsibilities on a single agent and abstracts away important aspects of real-world software development, including role specialization, peer review, and structured collaboration.

Recent work has explored increasing agent autonomy through richer action spaces and adaptive behavior. Self-evolving agents dynamically extend their toolsets during task execution \citep{xia2025livesweagent}, while generalist agent platforms provide flexible runtimes for executing complex sequences of actions across diverse tasks \citep{wang2024openhands}. However, these systems largely preserve a single-agent or loosely delegated execution model, and are often designed or tuned explicitly around benchmark settings. As a result, it remains unclear how well such systems transfer to real-world development workflows that prioritize reliability, traceability, and cost efficiency over benchmark-specific optimization.

In practice, software engineering is organized as a team-based process. Distinct roles such as project coordination, implementation, review, and research interact through shared artifacts and communication channels. On platforms like GitHub, this structure is formalized through issues, pull requests, inline code reviews, and iterative feedback loops. Since SWE-bench is derived from real GitHub issues, its tasks implicitly assume such workflows. Modeling software engineering as an isolated patch-generation problem therefore omits an important dimension of realism.

Beyond workflow fidelity, organizational decomposition enables separation of responsibilities, which is critical for scalable and efficient agent design. Different stages of software engineering impose different requirements on context size, reasoning depth, tooling, and model capabilities. Exploratory tasks such as issue analysis, repository understanding, and solution planning benefit from larger context windows and stronger reasoning, while implementation and debugging require precise code editing and rapid feedback cycles. A single-agent setup forces these heterogeneous requirements into a uniform configuration, often leading to suboptimal performance or unnecessary computational cost.

In this work, we present a fully automated multi-agent system built on top of \textsc{agyn}, an open-source platform for configuring agent teams with explicit communication, role-specific prompts, tool access, execution contexts, and LLM call tracing.

Importantly, this system was designed and deployed to solve real software development tasks in a production environment, where it is used as part of day-to-day engineering workflows. The system was not created specifically for SWE-bench, nor were its agents, prompts, or models tuned for the benchmark. Instead, SWE-bench is used as a post hoc evaluation to assess how a production-oriented, benchmark-agnostic system performs on a standardized academic benchmark.

Rather than treating software engineering as a monolithic task, our system decomposes it into specialized responsibilities carried out by a coordinated team of agents. We configure a multi-agent team consisting of a manager, engineer, reviewer, and researcher. Each agent operates with a well-defined objective, its own workspace, and a tailored set of tools. Agents may be backed by different language models depending on their role. In our setup, implementation-focused agents use smaller, code-specialized models such as GPT-5-Codex to efficiently perform coding and debugging tasks, while agents responsible for issue analysis, repository exploration, and task specification use larger general-purpose models such as GPT-5 with stronger reasoning and broader context capabilities. This role-specific model allocation reflects real production constraints and enables the system to balance solution quality, cost, and execution efficiency.

To mirror real-world development practices, the system executes a GitHub-native workflow that closely resembles human collaboration. Issues are analyzed, changes are proposed through pull requests, feedback is provided via inline code reviews, and revisions are iteratively applied until the solution is accepted. The entire process is fully automated and does not involve any human intervention. To support this workflow, we develop custom tools that provide agents with fine-grained access to GitHub primitives, including automated pull request creation, submission, and inline review capabilities.

We evaluate this production-deployed system on SWE-bench 500 under a fully automated setting. Our results demonstrate that modeling software engineering as a multi-agent organizational process, combined with native GitHub interaction and role-specific model configurations, is a viable and competitive approach to autonomous issue resolution, even without benchmark-specific tuning. Beyond benchmark performance, our work highlights the importance of workflow design, responsibility separation, and organizational structure as first-class considerations in the development of autonomous software engineering systems.

Our central hypothesis is that explicitly modeling software engineering as an organizational process with role and workspace separation, coordination, and review can be as important for autonomous issue resolution as improvements in underlying agents.

\paragraph{Contributions.}

\begin{itemize}[leftmargin=*, itemsep=2pt]
\item An open-source platform (\texttt{agyn}\footnote{\url{https://github.com/agynio/platform}}) for configuring and orchestrating multi-agent systems with explicit communication, roles, and dedicated sandboxes for autonomous research and execution.
\item A structured multi-agent software engineering team with distinct roles, prompts, tools, and role-specific model configurations.
\item Custom tooling that enables agents to perform inline code reviews and manage pull requests autonomously.
\item An empirical evaluation on SWE-bench 500, demonstrating competitive performance, with a 7.4\% higher issue resolution than the baseline mini-SWE-agent under comparable model configurations.
\item An open artifact set: forked SWE-bench repositories, opened issues and pull requests, and visible agent communication traces.
\end{itemize}

\section{Background and Related Work}

SWE-bench is a benchmark designed to evaluate autonomous software engineering systems on real-world tasks derived from GitHub repositories \citep{jimenez2024swebench}. Each task consists of a natural language issue description, a snapshot of the repository state, and a test suite used for verification. Successfully resolving a task requires understanding the issue, navigating the codebase, producing code changes, and validating the solution through testing. As such, SWE-bench evaluates end-to-end software engineering behavior rather than isolated code generation.

Early approaches to SWE-bench primarily relied on single-agent systems augmented with execution tools such as shell access, code editing, and search \citep{yang2024sweagent}. These systems embed a language model in an interaction loop that alternates between reasoning and acting within a repository environment. While this approach demonstrated that LLM-based agents can resolve a subset of real issues, it also exposed limitations related to long-horizon reasoning, error recovery, and the difficulty of handling heterogeneous subtasks within a single agent configuration.

A representative lightweight single-agent baseline is mini-SWE-agent \citep{minisweagent}, which deliberately minimizes agent scaffolding in order to emphasize language model capabilities. The system relies solely on Bash execution, maintains a linear interaction history, and executes actions as independent subprocess calls, making it easy to sandbox and evaluate across different models. This design makes mini-SWE-agent well suited as a baseline for SWE-bench, particularly in settings that aim to isolate model performance from agent infrastructure.

A class of approaches focuses on providing general-purpose execution environments for software engineering agents. A representative example is OpenHands \citep{wang2024openhands}, which offers a sandboxed runtime in which agents can modify code, execute commands, browse the web, and interact with external tools. These platforms aim to decouple agent logic from low-level execution details by exposing standardized interaction primitives and isolated execution environments. Such environments allow agents to iteratively modify code, run tests, and inspect results during task execution.

Another line of work explores increasing agent capability through adaptive behavior during task execution, including systems that dynamically synthesize new tools or utilities in response to task demands \citep{xia2025livesweagent}. By expanding the agent’s action space beyond a fixed scaffold, these approaches improve flexibility and have demonstrated strong performance on SWE-bench variants. However, they typically retain a monolithic execution model, where a single agent remains responsible for the full task lifecycle.

Multi-agent architectures have also been explored for software engineering tasks in the form of pipeline-based systems. TRAE organizes agents into a fixed sequence for patch generation, pruning, and selection, with different agents potentially using different configurations tailored to their roles \citep{gao2025trae}. Prometheus similarly structures issue resolution as a predefined sequence of agents for classification, issue reproduction, patch generation, and verification, supported by structured repository retrieval \citep{chen2025prometheus}. While these systems differ in emphasis (TRAE focusing on candidate generation and selection, and Prometheus on structured context retrieval and verification), they share a common design choice of encoding software engineering as a predefined multi-stage pipeline, where the number of stages and interactions is fixed in advance.

Across these approaches, most systems are designed and evaluated primarily with benchmark performance in mind. Prompts, tools, and configurations are often refined based on evaluation feedback, and success is measured by resolution rates on standardized test sets. While this methodology enables controlled comparison, it provides limited insight into how such systems transfer to production software development environments, where constraints related to cost efficiency, traceability, reliability, and failure recovery are central.

\paragraph{Positioning of our work.}
Our approach differs from prior work by modeling software engineering as an open-ended, manager-coordinated process carried out by a team of specialized agents. Rather than defining a fixed execution pipeline or relying on large-scale parallel generation, we organize agents into roles with distinct responsibilities and allow the coordination process to evolve dynamically. The number of interaction steps, revisions, and review cycles is not predetermined and is instead governed by a manager agent based on intermediate outcomes.

Each agent in our system operates within its own execution environment, where it can independently modify code, run tests, and explore alternative solutions. These environments are treated as first-class components of the system and are aligned with agent roles, enabling isolation of experiments, parallel exploration, and controlled integration of changes. This mirrors how human developers work with local development environments while coordinating through shared artifacts.

Instead of encoding a rigid pipeline, we provide the manager agent with a methodology that reflects real software development practice: research and understanding, task specification, issue formulation, implementation via pull requests, and iterative review cycles until approval. This methodology constrains how work is organized without prescribing how many steps must be taken or how agents must interact at each stage.

Finally, our system was developed and deployed to solve real software development tasks in a production environment and is used as part of day-to-day engineering workflows. It was not designed specifically for SWE-bench, nor were its agent configurations or models tuned for the benchmark. SWE-bench is used as a post hoc evaluation to assess how a production-first, benchmark-agnostic system performs on a standardized benchmark.

Our objective is to build an autonomous software development team composed of AI agents, provide this team with the tools and environments required for coding, research, and testing, and study how such a team collaborates to resolve real software issues. This framing emphasizes organizational structure, coordination, and system-level behavior over isolated agent performance.

\section{Solution Overview}

Our solution is built on top of \textsc{agyn}, an open-source platform designed to construct and configure multi-agent systems operating over real software artifacts. The platform provides abstractions for agents, execution environments, coordination, and tooling. In this section, we describe the system design used in our evaluation, including agent roles, coordination logic, execution environments, automation-oriented interaction design, and the GitHub-native workflow.

At a high level, the system models software engineering as a team-based process carried out by multiple autonomous agents with distinct responsibilities. Agents are configured independently and do not share a single global context or prompt. Each agent operates within its own execution environment, while coordination is handled explicitly by a manager agent. The system does not follow a fixed pipeline or assume a predetermined number of steps; instead, the interaction pattern emerges dynamically based on task complexity and intermediate results.

\subsection{Agents and role specialization}

Agents are first-class entities in the system. Each agent is defined by a role-specific objective, a prompt encoding its responsibilities, a set of permitted tools, an isolated execution environment, and a language model configuration. This separation allows each agent to maintain a focused context aligned with its role and enables differentiated allocation of computational resources.

We configure a team consisting of four roles: manager, researcher, engineer, and reviewer.

The manager agent is responsible for overall coordination and process control. It synthesizes outputs from other agents, maintains the task state, and decides which agent to invoke next. Rather than executing work directly, the manager applies a high-level methodology inspired by real software development practice, including deciding when additional research is required, when a task specification should be refined, when implementation should begin, and when review or revision is necessary.

The researcher agent focuses on understanding the issue and the repository. It explores relevant code paths, identifies likely root causes, and produces structured task specifications that guide implementation.

The engineer agent is responsible for implementation. It modifies code, runs tests, and iteratively refines solutions within its execution environment. Implementation-focused agents use smaller, code-specialized language models to balance cost and effectiveness during iterative development.

The reviewer agent is responsible for evaluating proposed changes through the pull request review process on GitHub. It inspects code diffs and opens an explicit pull request review. If issues are detected, the reviewer leaves inline comments and requests changes; if the solution satisfies the task requirements and passes validation, the reviewer approves the pull request. This explicit approve-or-request-changes mechanism serves as a concrete acceptance signal for the system and is used to determine when implementation work can be considered complete.

Different agents may use different language models depending on their role. Reasoning-heavy agents such as the manager and researcher are backed by larger general-purpose models, while implementation-focused agents use smaller, code-specialized models. This role-specific configuration reflects production constraints and improves overall efficiency.

\subsection{Execution environments}

Each agent operates within its own isolated execution environment that supports code modification, command execution, and test running. These environments are treated as first-class components of the system and are aligned with agent roles. Agents can freely explore the codebase, test partial solutions, and discard failed attempts without affecting other agents or the shared repository state.

Agents are granted direct access to a shell within their environments. We intentionally avoid providing a large set of predefined tools or preconfigured dependencies. Beyond basic version control utilities (\texttt{git}) and GitHub interaction (\texttt{gh}), agents are given access to the Nix package manager, allowing them to install and configure project-specific dependencies as needed.

We found that starting from an empty or minimally provisioned environment is more effective than using predefined environments. In several tasks, particularly in benchmark settings, correct execution required adjusting dependency versions, such as downgrading the Python interpreter or specific libraries to match repository constraints. Preconfigured environments often constrained this flexibility and introduced implicit assumptions that conflicted with project requirements.

To further control context growth during execution, we modify shell behavior to handle large outputs explicitly. When a command produces output exceeding a predefined threshold (50{,}000 tokens), the output is automatically redirected to a file in a temporary directory rather than being returned inline. The agent receives a reference to this file and can selectively inspect its contents using subsequent shell commands. This mechanism prevents large command outputs from overwhelming the language model context while preserving full access to the information when needed.

By allowing agents to construct their environments explicitly through shell commands and package management, and by controlling how execution outputs are surfaced, the system better supports heterogeneous repositories and long-running tasks without unnecessary context expansion.

\subsection{Communication and coordination}

Agents in our system do not communicate directly with each other. Instead, coordination is explicitly mediated by the manager agent through a dedicated \texttt{manage} tool. This tool allows the manager to invoke other agents, provide them with task-specific context, and collect their outputs. All inter-agent communication therefore flows through the manager, which serves as the single coordination point. 

The manager agent applies a high-level development methodology that mirrors real software engineering practice. Tasks are organized around research and understanding, task specification and issue formulation, implementation via a pull request, and iterative review cycles until approval. This methodology defines how work is structured, while the number of iterations and transitions between phases are determined dynamically based on intermediate outcomes.

While \textsc{agyn} can support more complex communication patterns, including direct agent-to-agent interactions, we intentionally adopt a manager-centric coordination model in this work. This design simplifies control flow, improves traceability, and aligns with a project-lead style of software development.

\subsection{GitHub-native workflow and tooling}

Coordination between agents is grounded in a GitHub-native workflow, which serves as the primary medium for persistent state and collaboration. To clearly distinguish agent actions, we create separate GitHub accounts for each agent role, allowing agents to appear as distinct contributors when opening issues, submitting pull requests, and leaving reviews.

Initial iterations relied on GitHub API access through an MCP-based integration. In practice, API responses included substantial auxiliary metadata, which inflated context size and degraded agent performance, particularly in review-heavy workflows. To reduce this overhead, we transitioned most interactions to the command-line utility \texttt{gh}, which provides more compact, human-oriented outputs that can be further filtered using standard shell utilities when needed.

While \texttt{gh} is effective for repository inspection and pull request management, it does not adequately support reading and authoring inline pull request reviews. To address this, we developed a custom extension, \texttt{gh-pr-review}\footnote{\url{https://github.com/agynio/gh-pr-review}}, which exposes a compact interface for accessing review threads and submitting inline comments. This enables agents to fully participate in iterative pull request review loops in a form suitable for language model reasoning.

\subsection{Test-driven execution strategy}

Our system follows a test-driven execution strategy for implementation. The engineer agent is instructed to begin work by running the existing test suite to establish a baseline and to re-run tests after each substantive code change. When the initial test state is clean and failures correspond directly to the reported issue, this approach provides a reliable feedback signal that guides incremental fixes and helps prevent regressions.

In practice, some SWE-bench tasks exhibit pre-existing test failures that are not directly related to the requested change. In such cases, a strict test-driven loop can introduce ambiguity, as the engineer agent may attempt to address unrelated failures and expand the scope of changes beyond the original issue. This can shift the agent’s trajectory away from the intended solution and negatively affect overall coordination. We discuss the implications of this behavior in Section~\ref{sec:limitations}.

\subsection{Automation-oriented interaction design}

Modern language models are primarily fine-tuned for conversational settings and often exhibit behaviors optimized for human-in-the-loop interaction, such as requesting approval to proceed, proposing plans and waiting for confirmation, or producing partial outputs while expecting further instructions. Such behaviors are incompatible with a fully automated system, where progress must be made without external feedback.

To address this, we design the system explicitly for automation-first execution. Agent prompts emphasize autonomous task completion and discourage requests for approval or confirmation. In addition, we enforce objective acceptance criteria at the system level. For example, a task is considered complete only when a pull request has been approved by the reviewer agent, rather than when an agent produces a natural language completion message.

Because only the manager agent is allowed to interact with the user, we further constrain its behavior to prevent premature termination. The manager is required to signal completion by invoking a dedicated \texttt{finish} tool. If the manager produces a non-functional output instead of a tool invocation, the system responds with an instruction emphasizing that work must continue autonomously. This ensures that the manager remains inside the coordination loop until completion conditions are satisfied.

At the same time, to preserve the ability to emit informational output without breaking automation, we introduce a separate \texttt{send\_message} tool. This tool allows agents to communicate status or intermediate information without signaling task completion. Separating communication from control flow helps prevent conversational behaviors from interfering with autonomous execution.

Finally, to support long-running tasks and maintain a bounded context window, we implement automatic context summarization. When the accumulated interaction history reaches a predefined threshold, earlier parts of the context are summarized and replaced with a compact representation that preserves key decisions, artifacts, and a status of a plan. This approach maintains a bounded context window while preserving critical task state needed for continued execution and decision-making.

\section{Experimental Setup and Results}

We evaluate our system on SWE-bench 500 under a fully automated setting. The evaluated system is identical to the one used in production and was not modified or tuned specifically for the benchmark. We focus on the text-only SWE-bench 500 setting \citep{jimenez2024swebench} and leave evaluation on multimodal software engineering tasks \citep{yang2025swebenchmultimodal} to future work.

\subsection{System configuration}

We use a fixed configuration across all tasks. The manager and researcher agents are backed by GPT-5 (medium reasoning), while the engineer and reviewer agents use GPT-5-Codex (medium reasoning). Each agent employs automatic context summarization with a fixed symbol budget to bound context growth during long-running tasks.

For each task, we fork the corresponding repository and create a dedicated base branch that serves as the execution entry point. Agents are required to open a pull request against this branch, with the task identifier included in the pull request title. A task is considered successfully completed only when the pull request is explicitly approved by the reviewer agent. All forks and pull requests generated during evaluation are publicly available, allowing external inspection of agent behavior and coordination across tasks.

\subsection{Evaluation protocol}

All tasks are executed end-to-end without any human intervention. Once a task is started, the manager agent coordinates research, task specification, implementation, and review according to the methodology described in Section~3. Iteration continues until the reviewer approves the pull request or a termination condition is reached. No manual prompt edits, corrective actions, or tool interventions are applied during execution.

\subsection{Results}
On SWE-bench 500, our system resolves 72.2\% of tasks in a fully automated setting. To disentangle the effect of system design from underlying model capability, we focus on a comparison among systems built on GPT-5–family models. As shown in Table~\ref{tab:swebench500_gpt5}, our manager-coordinated, workflow-oriented multi-agent system achieves the highest resolution rate within this subset, outperforming single-agent and pipeline-based approaches using comparable model configurations.

\begin{table}[t]
  \caption{SWE-bench 500 results for systems using GPT-5–family models.}
  \label{tab:swebench500_gpt5}
  \centering
  \begin{tabular}{lr}
    \toprule
    System & Score (\%) \\
    \midrule
    \textbf{agyn + GPT-5 / GPT-5-codex (medium reasoning)} & \textbf{72.2} \\
    OpenHands + GPT-5 & 71.8 \\
    mini-SWE-agent + GPT-5.2 (high reasoning) & 71.8 \\
    mini-SWE-agent + GPT-5 (medium reasoning) & 65.0 \\
    \bottomrule
  \end{tabular}
\end{table}

While the comparison isolates systems using GPT-5–family models, higher absolute scores have been reported by systems built on more powerful models. For example, mini-SWE-agent with Gemini 3 Pro Preview achieves 74.2\% on SWE-bench 500. At the same time, within the broader leaderboard, many modern systems cluster tightly in the 70–75\% range despite employing substantially different architectures, coordination strategies, and underlying models. In this context, our result demonstrates that a manager-coordinated, workflow-oriented multi-agent system can achieve competitive performance without relying on multi-attempt candidate generation or benchmark-specific optimization, highlighting the impact of organizational structure and coordination alongside model choice.

\subsection{Discussion}

Small numerical differences on the leaderboard should therefore be interpreted with caution. Benchmark resolution rates capture only a subset of factors that determine real-world effectiveness, omitting considerations such as environment robustness, dependency management, automation stability, cost, and traceability. From this perspective, our results suggest that system-level design choices can meaningfully influence performance, even when using comparable or weaker model configurations.

Beyond aggregate performance, the central contribution of this work is demonstrating that team-based coordination among specialized agents is a viable and effective paradigm for autonomous software engineering. Rather than relying on a single monolithic agent or a rigid execution pipeline, our system models development as a collaborative process involving agents with distinct roles, responsibilities, and execution environments. The ability of agents to communicate through structured coordination, iterate through review cycles, and jointly converge on a solution is a key factor in the observed performance.

We believe this team-oriented approach represents a promising direction for future agentic systems. As the community increasingly focuses on developing strong, reusable agent skills (such as proficiency in specific programming languages, frameworks, and ecosystems, as well as skills traditionally associated with quality assurance, testing, debugging, and code review), we expect agents to become more specialized and effective within well-defined domains. The ability to organize agents with complementary skills into coherent teams, providing them with appropriate environments, coordination mechanisms, and communication primitives, has the potential to significantly boost overall system performance beyond what can be achieved by individual agents alone. From this perspective, progress in agent infrastructure and organizational design may play a role as important as improvements in underlying models.

\section{Challenges, Limitations, and Future Work}
\label{sec:limitations}

While our system achieves strong performance on SWE-bench 500, the evaluation revealed limitations that are largely induced by the benchmark setup itself. Agent performance is highly trajectory-dependent: early obstacles—such as failures during environment setup, dependency installation, or test execution—can significantly affect downstream behavior and solution quality. Although SWE-bench is derived from real GitHub issues, many tasks originate from repositories and issues created years ago, under assumptions about tooling and dependencies that no longer hold.

As a result, agents may encounter errors that are incidental to the task rather than indicative of incorrect reasoning or implementation. In some cases, these failures interrupted execution or caused the system to diverge irrecoverably. To remain close to real-world development conditions, where work may be interrupted and resumed, we allowed the system to continue from intermediate states when such failures occurred. This continuation was performed from persisted GitHub artifacts without modifying agent prompts, tool configurations, or intermediate outputs.For example, if an agent successfully created an issue or partial implementation but execution was interrupted before completion, we resumed the run from that state and instructed the system to proceed. This approach reflects practical engineering workflows and avoids conflating benchmark-induced infrastructure failures with agent capability.

\subsection{Testing, legacy dependencies, and infrastructure drift}

Our system follows a test-driven workflow in which the engineer agent runs tests before and after making changes. This approach works well when dependencies install cleanly and the initial test state reflects the intended repository behavior. However, for a substantial portion of SWE-bench tasks, these assumptions do not hold.

Although SWE-bench is derived from real GitHub issues, many tasks originate from repositories created years ago and depend on legacy language versions, libraries, or CI infrastructure that no longer function reliably in modern environments. As a result, agents may encounter failures unrelated to the task itself. For example, some repositories fail CI due to deprecated GitHub Actions components, producing errors such as: ``This request has been automatically failed because it uses a deprecated version of \texttt{actions/cache: v2}.''

When faced with such issues, agents attempt realistic recovery strategies, including dependency downgrades, environment reconfiguration, or CI adjustments. While appropriate in production settings, these efforts consume context, time, and resources, and can divert agents away from the original task. In several cases, early environment-related failures dominated the execution trajectory and prevented agents from returning to the intended fix.

This behavior reflects a mismatch between production-oriented development workflows and benchmark evaluations, where expected solutions may implicitly assume an idealized or historically accurate execution context. When we observed that such infrastructure-induced failures consistently disrupted execution, we explicitly instructed agents to deprioritize CI checks that were clearly unrelated to the requested code change, even though this differs from standard real-world development practice.

\subsection{Infrastructure and resource constraints}

Our evaluation was conducted on a single MacBook Pro while allowing multiple tasks to execute in parallel. Each task involved multiple agents operating in isolated execution environments and frequently running test suites inside Docker containers. Under these conditions, we encountered memory pressure and occasional out-of-memory (OOM) failures when parallelism exceeded available resources.

Rather than indicating a limitation of the approach itself, this experience highlights the opportunity enabled by modern agentic systems to tackle large volumes of software engineering tasks concurrently. Scaling such systems with resource-aware scheduling, adaptive concurrency control, and distributed infrastructure would enable autonomous agent teams to address large backlogs of real-world engineering tasks in parallel, potentially changing the paradigm of how software products are built and maintained.

\subsection{Over-engineering relative to benchmark targets}

Another recurring challenge is that agents often attempt to do more than what is expected by the benchmark. In real-world development, it is common to improve test coverage, address linting issues, or refactor related code as part of a fix. However, SWE-bench gold patches are sometimes intentionally minimal and may omit tests or tolerate stylistic issues.

As a result, patches produced by our system occasionally differed from the benchmark’s expected solution, even when they were functionally correct. This mismatch illustrates a tension between production-oriented development practices and benchmark evaluation criteria, and suggests that success on benchmarks may sometimes reward under-specification rather than robustness.

\subsection{Long-running tests and execution limits}

Some repositories include test suites that are computationally expensive or slow to execute. Because tool calls are subject to execution time limits, long-running tests were occasionally terminated prematurely. In response, agents attempted alternative strategies such as running subsets of tests, batching execution, or parallelizing test runs.

While often reasonable, these explorations introduced additional complexity and sometimes influenced agent trajectories in unintended ways. This highlights the need for better mechanisms to reason about test selection, time budgeting, and partial validation in autonomous systems.

\subsection{Coordination and communication semantics}

Effective coordination between agents proved to be non-trivial. In particular, it was initially difficult to ensure that agents followed a consistent communication protocol. At times, the manager agent waited for actions from other agents that had already completed their work, leading to stalled execution.

To address this, we explicitly required agents to signal completion and readiness for further tasks, and reinforced coordination semantics through the \texttt{manage} tool with clear instructions governing communication flow. After these adjustments, coordination behavior became significantly more reliable. This experience highlights that clear communication contracts are essential for stable multi-agent collaboration.

\subsection{Future directions}

These challenges point to several directions for future work. Improving robustness to early failures, introducing mechanisms for trajectory correction and rollback, and developing better abstractions for environment and dependency management remain open problems. In addition, more expressive coordination primitives and resource-aware execution strategies will be necessary as agent teams grow larger and more specialized.

More broadly, our experience suggests that progress in autonomous software engineering will depend not only on improvements in language models, but also on advances in agent infrastructure, communication design, and organizational structure. Enabling teams of specialized agents to collaborate effectively over long horizons remains a key challenge and a promising opportunity for future agentic systems.

\bibliographystyle{plainnat}
\bibliography{refs}

\end{document}